\documentclass[10pt,journal,compsoc]{IEEEtran}

\usepackage[utf8]{inputenc} % allow utf-8 input
\usepackage[T1]{fontenc}    % use 8-bit T1 fonts
\usepackage{url}            % simple URL typesetting
\usepackage{booktabs}       % professional-quality tables
\usepackage{amsfonts}       % blackboard math symbols
\usepackage{nicefrac}       % compact symbols for 1/2, etc.
\usepackage{microtype}      % microtypography

\usepackage{times}
\usepackage{epsfig}
\usepackage{graphicx} % more modern
\usepackage{subfigure} 
\usepackage{graphicx}
\usepackage{amsmath}
\usepackage{amssymb}
\usepackage{epstopdf}
\usepackage{enumitem}
\usepackage{rotating}
\usepackage{forloop}
\usepackage{color}
\usepackage{todonotes}
\usepackage{etoolbox}
\usepackage{algorithm}
\usepackage[algoruled,boxed,lined,algo2e]{algorithm2e}
\SetAlgorithmName{Alg.}{algorithmautorefname}{list of algorithms name}
\ifCLASSOPTIONcompsoc
  \usepackage[nocompress]{cite}
\else
  \usepackage{cite}
\fi

\def\bigO2{\mbox{${\cal O}$}}
\def\bigO{O}

\def\mL{\mathcal{L}}

\def\1n{\mathbf{1}_n}
\def\0{\mathbf{0}}
\def\1{\mathbf{1}}

\def\bphi{\boldsymbol{\phi}}

\def\bmu{\boldsymbol{\mu}}

\def\btheta{\boldsymbol{\theta}}

\def\btheta{\mbox{\boldmath{$\theta$}}}
\def\bTheta{\mbox{\boldmath{$\Theta$}}}

\def\I{{\bf I}}

\def\X{{\bf X}}

\def\a{{\bf a}}

\def\x{{\bf x}}

\def\a1{\mbox{\bf a}_1}
\def\a2{\mbox{\bf a}_2}
\def\a3{\mbox{\bf a}_3}
\def\a4{\mbox{\bf a}_4}

\def\btheta{\boldsymbol{\theta}}

 \newcommand{\ie}{\emph{i.e.\;}}
 \newcommand{\eg}{\emph{e.g.\;}}
 \newcommand{\etal}{\emph{et al.\;}}

\begin{document}
\title{Herding Generalizes Diverse $M$-Best Solutions}

\author{Ece Ozkan, %~\IEEEmembership{Student Member,~IEEE,}
        Gemma Roig,
        Orcun Goksel, % ~\IEEEmembership{Member,~IEEE,}
        and~Xavier~Boix% <-this % stops a space
\IEEEcompsocitemizethanks{\IEEEcompsocthanksitem E. Ozkan and O. Goksel are with the Department
of Information Technology and Electrical Engineering, Computer-assisted Applications in Medicine, ETH Zurich, Zurich, Switzerland.\protect\\
E-mail: eoezkan@vision.ee.ethz.ch
\IEEEcompsocthanksitem G. Roig and X. Boix are with the Center for Brains, Minds and Machines, Istituto Italiano di Tecnologia@MIT, Massachusetts Institute of Technology, Cambridge, 02139 MA.}}% <-this % stops an unwanted space
\IEEEtitleabstractindextext{%
\begin{abstract}
We show that the algorithm to extract diverse $M$-solutions from a Conditional Random Field (called divMbest~\cite{batra2012diverse}) takes exactly the form of a Herding procedure~\cite{Welling09}, \ie a deterministic dynamical system that produces a sequence of hypotheses that respect a set of observed moment constraints. This generalization enables us to invoke properties of Herding that show that divMbest enforces implausible constraints which may yield wrong assumptions for some problem settings. Our experiments in semantic segmentation demonstrate that seeing divMbest as an instance of Herding leads to better alternatives for the implausible constraints of divMbest.
\end{abstract}

}

\maketitle

\IEEEdisplaynontitleabstractindextext

% \fi
\IEEEpeerreviewmaketitle

\IEEEraisesectionheading{\section{Introduction}}

\IEEEPARstart{C}{onditional} Random Fields (CRFs) are probabilistic graphical models for structured prediction~\cite{Lafferty01}, which have been successfully  applied in  Computational Biology, Computer Vision, and  Language Processing. Most CRF-based methods deliver maximum a posteriori (MAP) of the CRF probability function, using an off-the-shelf algorithms that delivers the MAP labeling in an efficient way, \emph{e.g.}~\cite{Frey97,Boykov04}.

Another strand of research exploits diverse and highly likely hypotheses extracted from a CRF, rather than only delivering the MAP labeling~\cite{batra2012diverse, chen2013computing, guzman2014efficiently}. Batra~\etal\cite{batra2012diverse} showed that in a pool of diverse and likely hypotheses extracted from a CRF, there may be some hypotheses with much higher accuracy than the MAP labeling. Also, applications that interact with a human or with other algorithms may benefit from delivering multiple hypotheses, since the human or the algorithms can select among them~\cite{Parkiccv11,yadollahpour2013discriminative,premachandranempirical}.

In this paper we introduce a more general view of the successful algorithm by Batra~\etal\cite{batra2012diverse} to extract the $M$-best diverse hypotheses from a CRF. We will refer to this method as {divMbest} in the rest of the paper.  DivMbest has been extensively analyzed in the literature, \eg\cite{chen2013computing, guzman2014efficiently,Prasad2014nips}, and it is representative of the state-of-the-art for extracting diverse hypotheses.

Our analysis reveals that divMbest is a particular case of the Herding procedure by Welling~\cite{Welling09}. Herding is a deterministic dynamical system that generates samples given a set of statistical moments~\cite{Welling09,welling2009herding,gelfand2010herding,chen2010parametric,Chen11,chen2012super,bach2012equivalence,uai12,aistats15,colt15}. DivMbest is exactly a particular case of Herding, with the parameters set in a specific way. Namely, divMbest sets the statistical moments to enforce that the set of delivered hypotheses have equiprobable marginals.

We show that these underlying constraints enforced by divMbest are rather implausible, since enforcing equiprobable marginals generates hypotheses with a proportion of labels independent on the input. In practice, divMbest alleviates this problem by using small update rates to avoid recovering the equiprobable marginals. This theoretical insight gained from the generalization allows for designing better constraints for extracting diverse hypotheses in applications, such as image segmentation.

We analyze several alternatives to the underlying constraints of divMbest, which overcome the problems of assuming equiprobable marginals. We demonstrate the capabilities of these constraints on VOC11 dataset~\cite{Everingham10}, in semantic segmentation, and also, in interactive image segmentation~\cite{Roig13,Maji14}, where the user provides the ground-truth labeling for a subset of regions in the image. The experiments show that the hypothesis selected by an oracle, is significantly more accurate when the new constraints are used, than when using divMBest to recover equiprobable marginals. 
\section{Diverse $M$-best Solutions}
\label{sec2}

This section revisits methods for extracting hypotheses  from a CRF, putting special emphasis in divMbest, which is the main focus of this paper.

{\bf Notation:  }
We use $\X=\{X_i\}$ to denote the set of random variables or nodes that represent the labeling of the entities in the CRF. Let $\mL$ be the set of discrete labels, and $\x=\{x_i\}$ an instance of $\X$, where $x_i \in \mL$.
We denote $P ( \x | {\btheta})$ as the probability density distribution of a labeling 
$\X=\x$ modeled with the CRF, in which $\btheta$  are the parameters of the model. We represent as $\x^{\star}$ the most probable labeling of  $P( \x | {\btheta} )$.  
$P ( \x | {\btheta} )$ can be written as a Gibbs distribution~\cite{Hammersley71}, \ie
$P ( \x | {\btheta} ) = (1/Z) \exp( E_{\btheta}(\x))$, in which $E_{\btheta}(\x)$ is the energy function and $Z$ is the partition function.

We express the energy function with the \emph{canonical over-complete representation}.
This is $E_{\btheta}(\x)=\btheta^T \bphi(\x)$, in which $\btheta$ are the parameters of the energy function. $\bphi(\x)$ is the vector of potentials, or the so-called sufficient statistics. We use $\bphi_u(\x)$ to denote unary potentials, and $\bphi_p(\x)$ for pairwise.  We use $\bphi_u(\x)=(\bphi_u(x_1),\ldots,\bphi_u(x_i),\ldots)^T$, where each component  $\bphi_u(x_i)$ is also a vector,
the elements of which correspond to an entry of the labels in $\mathcal{L}$, \ie $\bphi_u(x_i) =(\phi_u^1(x_i), \dots,\phi_u^l(x_i),\dots\phi_u^{|\mathcal{L}|}(x_i))^T$, and $|\mathcal{L}|$ is the cardinality of $\mL$. Each entry of the vector $\bphi_u(\x)$ is equal to $\phi_u^l(x_i) = \I[x_i = l]$, where $\I[a]$ is an indicator function that is $1$ if $a$ is true and $0$ otherwise. Note that only a single entry in $\bphi_u(x_i)$ will be equal to $1$.  The pairwise potentials, $\bphi_p$, use an \emph{a priori} assumption that depends on the application, and we describe them below. Methods that deliver an estimate of the most probable labeling infer the maximum a posteriori (MAP) of $P( \x | {\btheta} )$, or equivalently, maximize the energy function, \ie~$\x^{\star}= \arg\max_{\x \in \mL^N} E_{\btheta}(\x)$.

{\bf Extract hypotheses from $P( \x | {\btheta} )$:  }
Approaches to extract hypotheses from $P( \x | {\btheta} )$ include the Monte-Carlo methods that extract samples from $P( \x | {\btheta} )$, \eg~\cite{Porway2011ClusterSampling, barbu2005generalizing}. These methods usually take long periods until the samples capture different modes of the distribution, that can be used as hypotheses.
There are also methods to directly extract the $M$-best MAP modes~
\cite{chen2013computing,weiss2003MBestLBP, fromer2009lp,batra2012efficient}.
Yet, the $M$-most probable modes may be very similar to each other, and hence, they may not represent a diverse set of hypotheses.

As a result, several authors stressed the need to incorporate diversity constraints~\cite{batra2012diverse, guzman2014efficiently,Prasad2014nips}. Among the algorithms that enforce diversity, DivMbest has gained a lot of popularity for its simple implementation and high effectiveness. In this paper, we analyze divMbest, which is introduced next.

{\bf {DivMbest} algorithm:  }
DivMbest is the relaxation of a MAP inference algorithm with constraints that guarantee diversity among hypotheses. Let $\x^{\star}_m$ be the hypothesis extracted at the $m$-th iteration of the algorithm.  The constraints that are added to the MAP inference algorithm, and later relaxed in practice, are the following: 
\begin{align}
\forall q < m \; : \;  \bphi_u(\x^{\star}_m)^T \bphi_u(\x^{\star}_q) < K,
\label{eqSimilarity}
\end{align}
 where $K$ is a parameter that controls the diversity between hypotheses.  Note that Eq.~\eqref{eqSimilarity} evaluates  the similarity between the mappings of the unary potentials, $\bphi_u(\x)$, by counting the number of nodes that take different labels.

To implement this in practice, divMbest relaxes the diversity constraints in Eq.~\eqref{eqSimilarity}, which allows for extracting a hypothesis  by running MAP inference only one time. Thus, divMbest does not use the constraint in Eq.~\eqref{eqSimilarity} in practice, but a procedure which is a relaxation of the constraint. The procedure consists  of  modifying the energy function at each iteration in order to reduce the probability of the hypotheses obtained previously. This is done by subtracting  $\lambda \bphi_u(\x_m^\star)$ from the parameters of the unary potentials, $\btheta_{u(m)}$, where $\lambda$ controls the diversity. Thus, divMbest does not guarantee that the similarity among $\x^{\star}_m$ is less than $K$, as in Eq.~\eqref{eqSimilarity}: DivMbest adjusts the  diversity of the hypotheses  through adding $\lambda\bphi_u(\x_m^\star)$ to the energy function. The divMbest algorithm is shown in Alg.~\ref{algsoftconst}, and this is the algorithm used in the literature and in the rest of the paper.

{\bf Beyond divMbest:}
Several improvements of divMbest have been recently introduced, which have surpassed the performance of divMbest in some  particular scenarios.

Guzman-Rivera~\etal~\cite{guzman2014efficiently,guzman2012multiple} introduce a method  that consists of learning multiple models
to generate one hypothesis from each model.
These models are jointly learned using a loss function to yield diverse and accurate hypotheses. This comes at the cost of increasing the training, since multiple models are learned.
In applications where training multiple models is feasible in practice (\eg memory and computational cost can be scaled), it is shown that extracting hypotheses from multiple models can obtain better accuracy than divMbest.

Prasad~\etal~\cite{Prasad2014nips} introduce more sophisticated diversity measures than the dot product introduced in Eq.~\eqref{eqSimilarity}, $\bphi_u(\x^{\star}_m)^T \bphi_u(\x^{\star}_q)$. It is shown that when the energy function is submodular, a minimization procedure can effectively extract hypotheses with the new diversity measures. In this case of submodularity, results demonstrate that the combination of several diversity measures leads to more accurate hypotheses.

Another strand of methods, presented by Kirillov~\etal~\cite{kirillov2015inferring}, introduces and exploits a new assumption, which is that the number of hypotheses that the user needs to extract is given in advance. This new assumption allows to pose the hypotheses extraction as a single MAP inference problem in which all hypotheses are jointly inferred. This allows for a significant improvement of the accuracy of the hypotheses extracted, as the inference problem can be tuned to the given number of hypotheses. To make the computational cost of the joint inference problem feasible in practice, several optimizations have been presented for submodular energy functions~\cite{kirillov2015m,Kirillov2016}.

Even though the aforementioned improvements of divMbest lead to a better accuracy of the hypotheses, these improvements constrain the applicability of divMbest in terms of using submodular energy functions, limiting the training time, or knowing the number of hypotheses in advance. In this paper, we analyze divMbest because to the best of our knowledge, it is the algorithm with the best performance  among the algorithms that do not exploit constraints that limit the applications. In contrast to previous works, the improvements of divMbest yield by our analysis do not constraint the applicability of divMbest.
\definecolor{mygray}{rgb}{1,1,1}
\definecolor{mygray2}{rgb}{1,1,1}
\definecolor{mygray3}{rgb}{1,1,1}

\begin{table*}[t]

\begin{minipage}[t]{0.45\linewidth\relax}
    \null 
  \begin{algorithm}[H]

\footnotesize
\KwIn{$\btheta$, $\lambda$}
\KwOut{$\{\x_m^\star\}$}

\vbox{\colorbox{mygray3}{\vbox{
\nl $\triangleright$ \emph{Initialization:} \\
$\btheta_{(0)} = \btheta$
}}}
\For{$m= 0: M$}{
\vbox{\colorbox{mygray}{\vbox{
\nl $\triangleright$ \emph{Maximization:} \\
$\x_m^\star = \arg\max_{\x} \btheta_{u (m)}^T \bphi_u(\x)  +\btheta_p^T \bphi_p(\x)$ 
}}}
\vbox{\colorbox{mygray2}{\vbox{
\nl$\triangleright$ \emph{Update:} \\
$\btheta_{u (m+1)} = \btheta_{u(m)}  - \lambda \bphi_u(\x_m^\star)$
}}}
}

\caption{{Diverse M-best Solutions}}
\label{algsoftconst}
\end{algorithm}
\end{minipage}\hspace{1.00cm}
\begin{minipage}[t]{0.45\linewidth\relax}
  \null
 \begin{algorithm}[H]
 \footnotesize
 \setcounter{AlgoLine}{0}
\KwIn{$\bmu$, $\eta$, $\bTheta$}
\KwOut{$\{\x_m^{\star}\}$}

\vbox{\colorbox{mygray3}{\vbox{
\nl $\triangleright$ \emph{Initialization:} \\
$\btheta_{(0)}=\bTheta$
}}}
\For{$m=0:M$}{
\vbox{\colorbox{mygray}{\vbox{
\nl $\triangleright$ 
\emph{Maximization:} \\
$\x_m^{\star} = \arg\max_{\x} \btheta_{(m)}^T \bphi(\x)$ 
}}}
\vbox{\colorbox{mygray2}{\vbox{
\nl $\triangleright$ 
\emph{ Update:}\\
$\btheta_{(m+1)}=\btheta_{(m)} + \eta(\bmu - \bphi(\x_m^{\star}))$
}}}
}

\caption{{ Herding}}
\label{AlgGrad}
\end{algorithm}
\end{minipage}%

\end{table*}

\section{Herding}

In this section, we revisit Herding, which is a deterministic dynamical system that generates samples that follow a given set of statistical moments. This will serve as the basis to show in the following section that {divMbest} is a particular instance of {Herding}. Herding has been extensively analyzed in the literature~\cite{Welling09,welling2009herding,chen2010parametric,Chen11,chen2012super}. Herding is related to the perceptron~\cite{gelfand2010herding}, to quadrature methods and Frank-Wolfe optimization methods~\cite{bach2012equivalence,uai12,aistats15}, and also, there is a recent analysis of Herding using discrepancy theory~\cite{colt15}. We refer to the references for further details. In this paper, we introduce a new relation between Herding and the extraction of diverse hypotheses.

{ \bf {Statistical Moments of the Gibbs distribution:}}
Before introducing Herding, we review Jaynes' celebrated result, that shows that the Gibbs distribution recovers a set of statistical moments, and it has maximum entropy~\cite{jaynes1957information}. This will serve as the basis to introduce Herding afterwards.

Let $\bmu$ be the vector of statistical moments that describe a set of samples $\{\x\}$, which 
we divide into unary and pairwise moments, denoted as $\bmu_u$ and $\bmu_p$, respectively. 
   $\bmu$ is computed by averaging the vector of sufficient statistics $\bphi(\x)$ over $\{\x\}$, \ie
\begin{align}
 \bmu = \mathbb{E}_{\x\sim f} [\bphi(\x)],
 \label{eqmu}
 \end{align}
where $f$ is the distribution that generated the samples $\{\x\}$.
 Note that since $\bphi(\x)$ describes the samples using indicator functions that take values $1$ or $0$, we can see that the moments $\bmu$ indicate the proportion of samples that each entry in $\bphi(\x)$ is equal to $1$. 
 The unary moments, $\bmu_u$, indicate the proportion of samples that $\x$ is equal to each label in the set of samples. Let $\bmu_u^i$ be an element of $\bmu_u$, which corresponds to the marginal distribution of $x_i$. Thus, it is constrained to sum up to $1$ ($\|\bmu_u^i\|_1=1$), where each entry in $\bmu_u^i$ is higher or equal to~$0$. Also, the elements of the pairwise moments, denoted as $\bmu_p^{ij}$,  correspond to the pairwise marginal distribution, and hence, the sum of the entries in $\bmu_p^{ij}$ is equal to $1$.
 Jaynes showed that the Gibbs distribution is the distribution with maximum entropy over all possible distributions that have moments $\bmu$. This is, Gibbs distribution is ``assumption''-free, beyond having statistical moments $\bmu$.
Jaynes' result follows from the solution to the following optimization problem:  
 \begin{align}
 \arg\max_{f} \tau \mathcal{H} (f), \;\;\mbox{s.t.} \;\bmu = \mathbb{E}_{\x\sim f} [\phi(\x)],
 \label{eqPrimal}
 \end{align}
where $\mathcal{H} (\cdot)$ is the Shannon entropy, and $\tau$ the temperature parameter. We can see in Eq.~\eqref{eqPrimal}, that the optimization over the probability distributions, $f$,  maximizes the entropy given the set of moments of the distribution, $\bmu$. Jaynes showed that the solution of this optimization problem is that $f$ is equal to the Gibbs distribution with energy function $E_{\btheta}(\x)=\btheta^T \bphi(\x)$, in which the parameters $\btheta$  are the Lagrange multipliers of the optimization problem in Eq.~\eqref{eqPrimal}.

Thus, the Gibbs distribution, or equivalently, the probability distribution of the CRF, is characterized either by the statistical moments, $\bmu$, or by the Lagrange multipliers of Eq.~\eqref{eqPrimal}, $\btheta$. In divMbest, the Gibbs distribution is characterized using $\btheta$.

{ \bf {Herding Reconstructs the Moments:}} 
Herding is a procedure to obtain samples with statistical moments $\bmu$. The underlying probability distribution of the samples obtained from Herding is unknown, although in some cases, it has been shown that the probability distribution of Herding is close to the Gibbs distribution with moments $\bmu$,~\ie the samples obtained from Herding might be close to the case of maximum entropy~\cite{bach2012equivalence}. The advantage of using Herding instead of sampling directly from the Gibbs distribution, is that the samples from Herding converge much faster to the statistical moments than the iid random samples from the Gibbs distribution ---$\bigO(M^{-1})$ \emph{vs} $\bigO(M^{-{1/2}})$, respectively.

The procedure of Herding to obtain a set of samples with statistical moments $\bmu$ is summarized in Alg.~\ref{AlgGrad}. Each iteration of the algorithm consists of two steps, which collectively generate one sample. First step is a maximization of an energy function, which is done with an off-the-shelf MAP inference algorithm. $\x^{\star}_m$ denotes the optimal value obtained from this maximization at the $m$-th iteration of Herding. Second step is the update of parameters of the energy function $\btheta_m$, which are used in the optimization in the first step. The update rate is denoted as $\eta$ and remains fixed at a constant value.
Herding initializes  $\btheta$ equal to $\bTheta$, which is an input value given to the algorithm.

Herding does not specify the analytical form of the probability distribution that generates $\{\x^{\star}_m\}$,~\ie it does not specify $f$. Yet, Herding generates samples $\{\x^{\star}_m\}$ that reconstruct the moments $\bmu$.
Namely, Welling~\cite{Welling09} showed that if $\eta>0$, $\|\bphi(\x)\|_2$ is bounded for all $\x$, and $\bmu$ is in the marginal polytope (\ie $\bmu$ is the average of the elements of any non empty set $\{\bphi(\x_i)\}$), then the samples generated from Herding greedily minimize the following reconstruction error of the moments: 
\begin{align}
\|\bmu - \frac{1}{m}\sum_{k=1}^m \bphi(\x^{\star}_k)\|_{2}^{2},
\label{eqHerdingRecovery}
\end{align}
and the minimization converges at a rate $\bigO(M^{-1})$. If $\bmu$ is not in the marginal polytope, Chen~\etal\cite{Chen11} showed that a simple normalization of $\btheta_{(m)}$ by a constant after each iteration of Herding (to prevent $\btheta_{(m)}$ to diverge), produces that the samples of Herding achieve the global minimum of Eq.~\eqref{eqHerdingRecovery} at the rate $\bigO(M^{-1})$.

Different learning rates $\eta$ could be used by different moments,~\eg the learning rate of the unary and the pairwise terms are different between them. Chen~\etal\cite{Chen11} showed that the convergence guarantees apply with the L2 distance in Eq.~\eqref{eqHerdingRecovery} weighted by the learning rates.  Also,  when the global maximum in MAP inference is not achieved, the convergence guarantee can hold true. It only requires that $\x^\star_m$ fulfills the following condition: $\btheta_{(m)}^T\bmu\leq \btheta^T_{(m)}\bphi(\x^\star_m)$~\cite{chen2012super}.  These results are remarkable, as they guarantee the moments of the samples from Herding quickly converge to $\bmu$ under very mild conditions.

{\bf Herding for Diverse Sampling: }
Herding can be rewritten in an equivalent form which intuitively shows that Herding generates diverse samples.
As shown by Chen~\etal\cite{chen2012super},  a new sample of Herding, $\x^\star_{m+1}$, is obtained by inferring the MAP labeling  of the following expression: 
\begin{align}
\arg\max_{\x\in\mathcal{L}^N} 
&  \underbrace{ \frac{1}{\eta m} \bphi(\x)^T \bTheta}_\text{Initialization dependent} -\underbrace{ \frac{1}{m}\sum_{k=1}^{m} \bphi(\x)^T \bphi(\x_k^\star)}_\text{Diverse Samples} +\underbrace{\bphi(\x)^T \bmu}_\text{Recover $\bmu$}.
\label{eqDiverse}
\end{align}

First term in Eq.~\eqref{eqDiverse} depends on the initialization of $\btheta$. Note that it becomes less influential after extracting several samples, as this term is divided by the iteration number, $m$.

Second term is the similarity between a new sample and the samples that have been generated previously. Thus, Herding encourages that the samples are as different as possible among them, which may help to quickly reconstruct the moments.  Note that the similarity used in Eq.~\eqref{eqDiverse} is the same as in Eq.~\eqref{eqSimilarity} of {divMbest} before the relaxation, with the difference that in Eq.~\eqref{eqDiverse} all the potentials are taken into account, whereas in Eq.~\eqref{eqSimilarity} only the unary potentials, $\bphi_u(\x)$, are used.

Third term in Eq.~\eqref{eqDiverse} encourages the correlation of $\bphi(\x)$ with the moments $\bmu$. As we mentioned before, there are guarantees the moments of the samples converge to $\bmu$. 

Finally, in Eq.~\eqref{eqDiverse}, we can see that there is not a parameter that directly controls  the diversity of the samples. Yet, note that the diversity comes from the moments $\bmu$ and $\eta$. The moments $\bmu$ specify the statistics of the samples, and hence, also the diversity. The parameter $\eta$ controls the influence of the initialization parameters. For small $\eta$, Herding encourages that the samples are more similar to $\bTheta$ than for high values of $\eta$.
\section{Herding Generalizes DivMbest}
\label{secGeneralization}

Now we show that {divMbest} is an instance of Herding. Observe that the differences between the algorithm of {divMbest} (Alg.~\ref{algsoftconst}) and Herding  (Alg.~\ref{AlgGrad}) are small. Recall that $\eta$ is the update rate in Herding. Let $\eta_u$  be the update rate of the unary terms, and $\eta_p$ for the pairwise terms. We can recover {divMbest} from Herding by setting the unary moments of Herding and the pairwise update rate to zero, and using $\btheta$ as the initial value of the dynamical system. Thus, divMbest is equal to Herding with the following parameters:
\begin{align}
  \bmu_u = \0,\;\;\; \eta_p = 0, \;\;\; \bTheta=\btheta.
\end{align}
 Observe that this allows to recover the update step of {divMbest} in Alg.~\ref{algsoftconst} from Herding in Alg.~\ref{AlgGrad}. Then, note that $\eta_u$ is the parameter $\lambda$, and the pairwise parameters $\btheta_p$ do not change in {divMbest} because $\eta_p=0$, and hence,  the pairwise moments, $\bmu_p$, do not need to be defined in divMbest.  The initialization of Herding is set as the initialization in {divMbest},~\ie $\bTheta$ is equal to the CRF potentials.

Thus, the formulation of {divMbest} as Herding gives an interpretation of the hypotheses delivered by {divMbest} as being generated by a deterministic dynamical system that is reconstructing unary moments equal to $\0$, \ie~$\bmu_u = \0$. This observation begs the question whether always enforcing $\bmu_u = \0$ independently on the input is a plausible constraint.

{\bf Consequences of $\bmu_u=\0$ in divMbest: }
For the sufficient statistics we use, the closest reconstructable moments to $\bmu_u = \0$ are the unary moments the labels of which have equiprobable probability of occurrence. This can be seen by substituting $\bmu_u = \0$  and $\eta_p=0$ in Eq.~\eqref{eqHerdingRecovery}, which yields $\|\frac{1}{M}\sum_{m=1}^M \bphi_u(\x^{\star}_m)\|_2^2$, and finding the set of samples that minimize this expression, in the same way as Herding. Since the pairwise terms have been cancelled ($\eta_p=0$), we can minimize Eq.~\eqref{eqHerdingRecovery} for each unary term independently, \ie~$\{\|\frac{1}{M}\sum_{m=1}^M \bphi_u^i(\x^{\star}_m)\|_2^2\}$.  Recall that $\frac{1}{M}\sum_{m=1}^M\bphi_u^i(\x^{\star}_m)$ lies in the marginal polytope,~\ie~$\|\frac{1}{M}\sum_{m=1}^M\bphi_u^i(\x^{\star}_m)\|_1=1$. Since Eq.~\eqref{eqHerdingRecovery} minimizes the $\ell_2$ norm, and the marginal polytope lies in the $\ell_1$ ball of radius $1$, we can see by simple geometry that the point in this $\ell_1$ ball closest  to $\0$ in terms of the $\ell_2$ distance is the point of  equiprobable unary marginals.

From these observations and the convergence guarantees of Herding introduced in the previous section, we conclude that the samples of {divMbest} converge to equiprobable unary marginals, independently of the parameters of the problem and the input.

{\bf How does {divMbest} alleviate enforcing implausible constraints ($\bmu_u=\0$)? }
The way that {divMbest} mitigates enforcing the moments $\bmu_u = \0$  is to use a slow update rate, $\eta_u$ (\ie the parameter called $\lambda$ in divMbest notation).  Recall the parameter that controls the diversity in Herding are $\eta$  and the moments $\bmu$. In divMbest, the moments are predefined to $\bmu_u = \0$, which is the maximum possible diversity as it encourages equiprobable distribution of labels. Recall that  $\eta$ weights the initialization term in Eq.~\eqref{eqDiverse}, $\frac{1}{\eta m} \bphi(\x)^T \bTheta$. Then, small values of $\eta$ encourage hypotheses similar to the initialization and reduce the influence of the attractor that yields $\bmu_u=\0$. In the experiments we show empirical evidence that support these theoretical observations.
\section{Experiments}

In this section,  we analyze divMbest with different constraints to extract multiple hypotheses for image segmentation.
Note that most methods to extract diverse hypotheses can also be used in applications different from image segmentation. Some examples of applications are camera pose estimation~\cite{guzman2014camera}, automatic text translations~\cite{gimpel2013systematic}, and body pose estimation~\cite{Park2011PartsNBest, batra2012diverse}. Also, in more general settings, the multiple hypotheses can be used for estimating the uncertainty of the MAP labeling~\cite{Ramakrishna2012nipsw} or  for training structural SVMs in an  efficient way~\cite{guzman2013divmcuts}. We have chosen image segmentation because it uses the probabilistic models in a simple way, and it serves to illustrate the theoretical findings and the importance of replacing $\bmu_u=\0$ by other constraints that are more plausible.

\subsection{Benchmarks}

We report results on VOC 2011~\cite{Everingham10} which 
has pixel-wise annotations of $20$ object classes plus background. We use VOC11 because there are publicly available pre-trained models~\cite{Carreira12}, which will allow the reproducibility of our experiments. We report results in the validation set, since the ground-truth for the testing set is not provided.

We use two different modalities of image segmentation: \emph{(i)}~{semantic image segmentation}, and \emph{(ii)} interactive image segmentation with missing potentials.
Semantic segmentation consists on assigning a labeling to each pixel corresponding to its semantic class, from a predefined set of semantic classes. The algorithm to automatically recognize each semantic class can be learned from a training set.
In the case of interactive image segmentation,  the user marks few pixels or superpixels providing their true semantic label, which are used to instantiate the corresponding unary terms.  The probabilistic inference should propagate the information in the known unary terms through the pairwise information. Since there is missing information, delivering several hypotheses of image labeling may help the user to pick the best segmentation hypothesis among all.

Some applications of interactive segmentation target to segment the foreground in one image, and not the object semantics as in our case~\cite{batra2012diverse,yadollahpour2013discriminative,premachandranempirical}. In the case of foreground segmentation, the unary terms can be estimated in the whole image by learning a  model of the foreground (usually based on color). Note that the interactive segmentation we are evaluating is more challenging than foreground segmentation because it is not possible to have an estimate of the unary potentials in all the image  (we can not train an object classifier from few superpixels given one image).

{\bf Evaluation metrics:} We use the standard evaluation metrics provided with VOC11. We evaluate the hypotheses using the criteria of \emph{mode} and \emph{oracle}. The {mode} is obtained by selecting the most frequent label among the hypotheses in each pixel, and gives an idea of the performance for the average of the hypotheses. Thus, it is useful to analyze the properties of the moments of the extracted hypotheses. The {oracle} selects the final labeling, \ie the hypothesis that is most similar to the oracle (ground-truth).  In all the plots the accuracy is the average accuracy over all classes.

\begin{figure}[t!]
\centering
\includegraphics[width=0.42\textwidth]{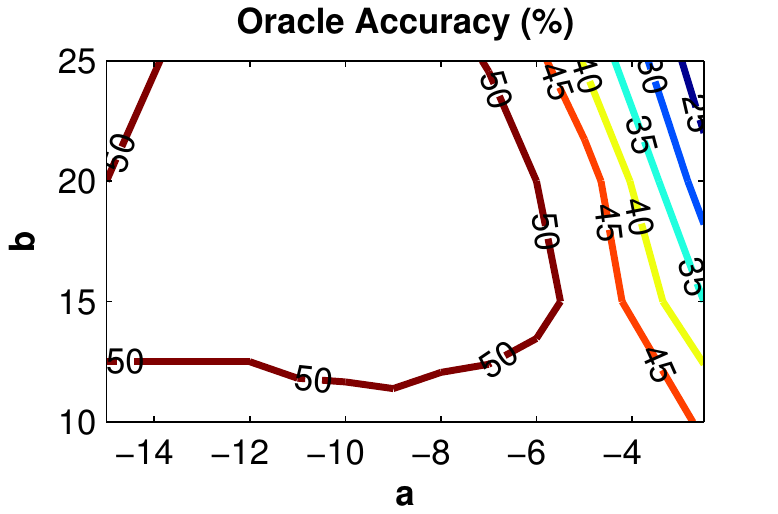} 
\caption{\emph{Impact of the sigmoid parameters in VOC11.}  Oracle accuracy of divMbest for $M=20$ is reported for different values of the sigmoid parameter.} 
\label{fig:sigmoid}
\end{figure}

\subsection{CRF for Image Segmentation}

In the two modalities of image segmentation that we report results, $\X=\{x_i\}$ represents a semantic labeling of the regions in the image, and $x_i$ takes a value from a predefined set of labels. We use the standard CRF for semantic segmentation that has as unary potentials the output of an object recognition algorithm, and it has as pairwise potentials the color modulated Potts model, which encourages the same labeling among neighboring regions with similar color~\cite{BoykovPotts2011}. The parameters that we report are set via two-fold cross-validation in the validation set of VOC11, by optimizing for the oracle accuracy for $M=20$ hypotheses. These parameters optimized for $M=20$ are used for extracting any number of hypotheses, as we do not assume we know in advance the number of hypotheses the user may need to extract. Next, we provide the implementation details. 

%More sophisticated constraints could also be incorporated, but these constraints have.

 {\bf Superpixels:} We use non-overlapping superpixels that are provided in CPMC by Carreira~\etal\cite{carreira2010constrained} for VOC11 (about $600$ per image). We define one random variable for each superpixel.

{\bf Pairwise Term: } The potential $\bphi_p(x_i,x_j)$ is based on the smoothness of the labeling, and it is the following vector: $\bphi_p(x_i,x_j) = (\I[x_i = x_j], \I[x_i \neq x_j])^T$, in which $i$ and $j$ index neighboring regions in the image. $\btheta_p^{ij}$ is initialized to the color-modulated Potts potentials of the CRF, \ie~$\btheta_p^{ij}=(0,-C)$, where $C$ is a constant that depends on the similarity between the regions. In this way, when the regions are similar ($C$ with a high value), the Potts potential encourages them to have the same label; and when the regions are dissimilar ($C$ with values close to $0$), this potential does not enforce any constraint. 
To evaluate the similarity between neighboring regions, we use  the exponential of the normalized Euclidean distance between the mean of the RGB values 
of the connected superpixels, with decreasing factor equal to $10$ for semantic segmentation, and equal to $1$ for interactive segmentation. Finally, in order to re-weight the pairwise term in the energy function, the pairwise potentials are multiplied by $0.08$ for semantic segmentation, and $0.15$ for the interactive segmentation, which are found by cross-validation. Note that the parameters for the interactive segmentation encourage more propagation of the labeling, as in this application the propagation is more important to obtain better accuracy because there are unknown unary terms.

{\bf Unary Term:}
For semantic segmentation, to obtain the object classification scores for each superpixel, we use the publicly available precomputed models by~\cite{Carreira12} in the VOC11. The unary potentials, $\btheta_{u(0)}$, are initialized to the negative logarithm of the scores' probability given by the classifiers (following the literature on divMbest). Namely, we use classification scores computed with second-order-pooling on the CPMC regions as in~\cite{Carreira12}, using the publicly available code. We select for each superpixel the CPMC region with the highest score, and use the scores in this region as the scores of the superpixel.  We use a sigmoid function to map the classification scores to probabilities, \ie~$\left(1+\exp(-(a+b\cdot s)) \right)^{-1}$, where $s$ is the output of the classifiers, and $a$ and $b$ are two parameters that we learn with cross-validation. In Fig.~\ref{fig:sigmoid}, we show the oracle accuracy performance of divMbest for $M=20$ for different values of $a$ and $b$. We can see that the performance is slightly affected by different choices of $a$ and $b$. The optimal values found by cross-validation are $a = -7$ and $b = 15$. We use these values in the rest of the experiments.
Finally, the output of the sigmoids are normalized to yield probabilities, and the negative logarithm of the probabilities are used as initialization of the unary potentials.

The unary potentials for the interactive image segmentation application correspond to the true annotation for the locations where the human has provided information. We assume that the user selects some random points of the image and assigns their ground-truth to their unary potentials. The rest  of the unary terms remain as missing or unknown~\cite{Roig13}.

 {\bf Inference of the Hypotheses: } We use loopy belief propagation by~\cite{Frey97} for MAP inference (it takes $10$ms to converge in one CPU of a MacBook Air). Note that graph cuts~\cite{Boykov04} can not be used due to the non-submodular functions that may arise during Herding updates.
The MAP labeling gives $44.8\%$ accuracy for the VOC11 validation set, which is similar to the one reported in~\cite{Carreira12}.

\begin{figure*}
\begin{tabular}{@{\hspace{1mm}}c@{\hspace{1mm}}c@{\hspace{1mm}}c}
\multicolumn{2}{c}{\hspace{-1mm}\includegraphics[width=0.6\textwidth]{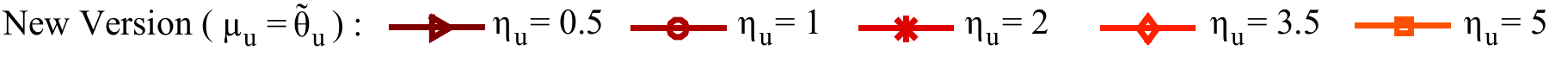}}&\includegraphics[width=0.18\textwidth]{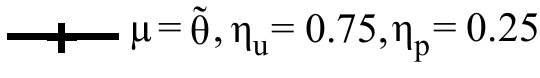}\\
\multicolumn{2}{c}{\hspace{-1mm}\includegraphics[width=0.6\textwidth]{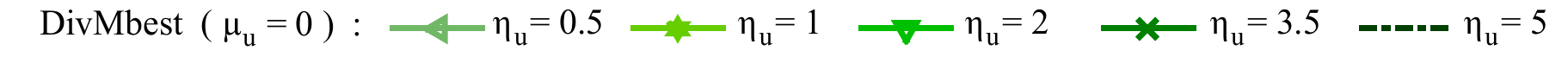}}& \includegraphics[width=0.25\textwidth]{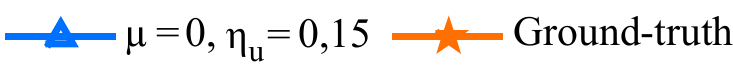}\\
\includegraphics[width= 0.315\textwidth]{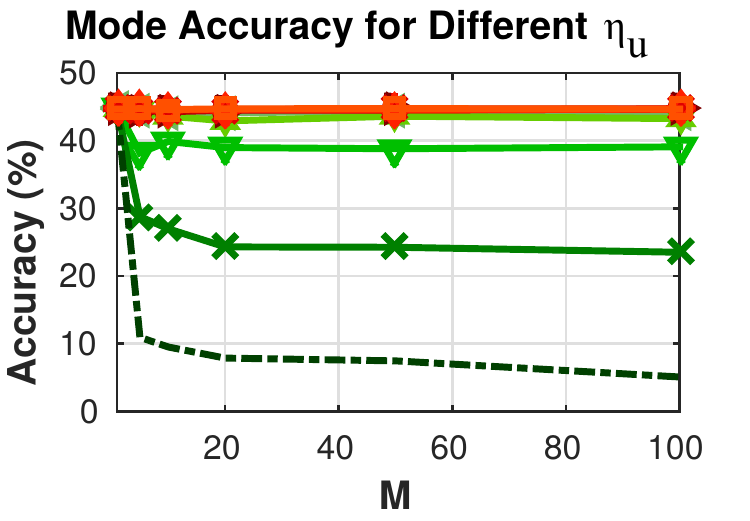} &
\includegraphics[width=0.31\textwidth]{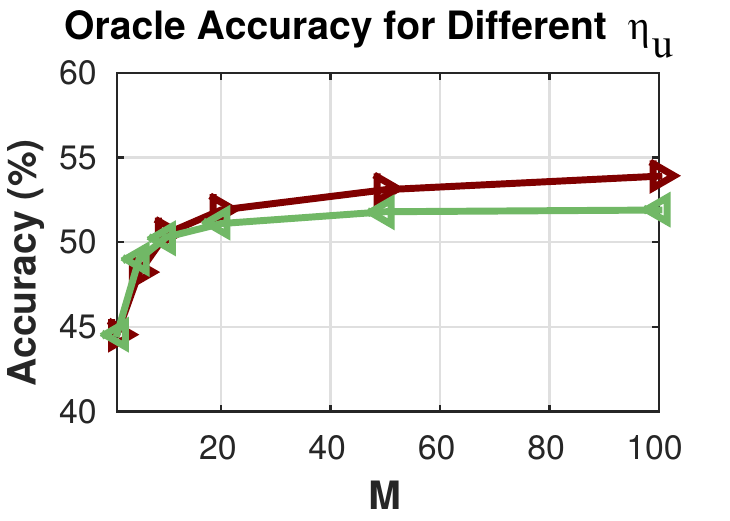} &
\includegraphics[width=0.33\textwidth]{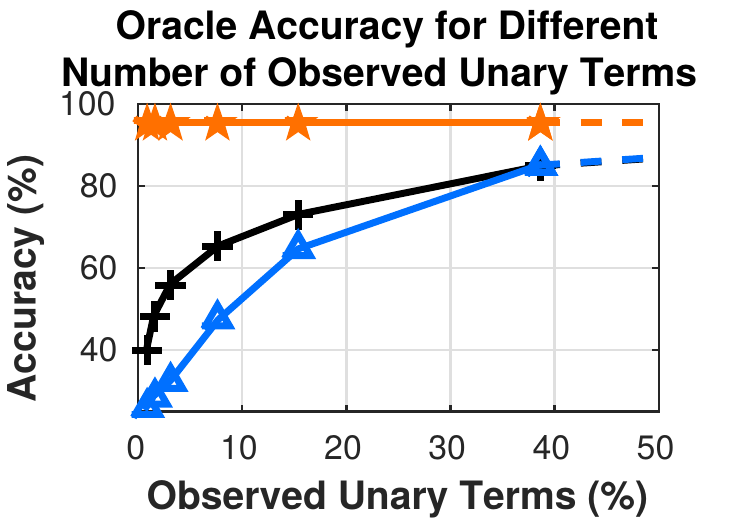} \\
{ (a)} & { (b)} & { (c)}    \\
\end{tabular}
\caption{\emph{Evaluation of divMbest in VOC11 for image segmentation.} In (a) we compare mode performance for divMbest and divMbest's new version with different $\eta_u$, and in (b) we report oracle performance for divMbest and its new version with optimal $\eta_u$. In (c), we evaluate the oracle performance of interactive image segmentation for different number of observed unary terms. Ground-truth is with $100\%$ of the unary moments observed.}
\label{fig:herdingWithVsWithoutExt}
\end{figure*}

\subsection{Results on Semantic Segmentation}
Recall that from the generalization, {divMbest} is a Herding procedure with $\bmu_u=\0$, $\bTheta=\btheta$, and $\bmu_p$ is not defined because the update rate $\eta_p$ is $0$.
As described in sec.~\ref{secGeneralization}, divMbest enforces implausible constraints to the unary moments, $\bmu_u=\0$, and  this can be alleviated by using a small update rate, which keeps the dynamical system close to the initialization.

In order to validate that the constraint $\bmu_u=\0$ is implausible because it enforces equiprobable marginals, we report experiments with a new version of divMbest that does not enforce equiprobable marginals. Note that we can not remove the constraint on the moments because it is necessary to extract hypotheses, but we can replace the constraint currently implemented in divMbest,  $\bmu_u=\0$, by another constraint more plausible.

 In the new version of divMbest, we replace $\bmu_u=\0$ by unary moments based on the parameters of the unary potentials of the CRF. Namely, we denote $\tilde{\btheta}_u$ as the parameters of the unary potentials, $\btheta_u$, converted to a vector of moments by inverting the negative logarithm in $\btheta_u$ and applying a normalization to make $\tilde{\btheta}_u$ represent a vector of marginal probabilities.  In the new version of divMbest we replace the equiprobable moments $\bmu_u=\0$ by $\bmu_u=\tilde{\btheta}_u$.   This constraint on the moments is more plausible than  $\bmu_u=\0$, because it enforces that the samples follow a set of moments proportional to the output given by the object recognition algorithm,~\ie an object category should appear in more samples if the output of the object recognition algorithm is high for this category. Note that this new version of divMbest does not give any advantage besides replacing the implausible constraint $\bmu_u=\0$, as we do not introduce any additional information,~\ie the unary moments are defined using the unary term of divMbest $\bmu_u=\tilde{\btheta}_u$. Also, in this new version of divMbest, the initialization is done as in divMbest, $\bTheta=\btheta$. In the figures and tables, we denote this version of divMbest  as $\bmu_u=\tilde{\btheta}_u$.

In Fig.~\ref{fig:herdingWithVsWithoutExt}a we show the mode accuracy, which allows a direct analysis of the constraint $\bmu_u=\0$. This is because the mode accuracy evaluates the unary marginals of the hypotheses. We report results for different update rates of the unary moments, $\lambda$ (or equivalently $\eta_u$), for both divMbest and the modified divMbest with the moments proportional to the unary potentials.
We can see that  when we increase the update rate,  $\lambda$ or $\eta_u$, the mode performance decreases for divMbest but it remains the same for the modified divMbest. This is because  divMbest is greedily enforcing $\bmu_u=\0$, which produces a drop of the mode accuracy as the average of the hypotheses becomes equiprobable. Observe that the mode accuracy sometimes enters in a plateau and does not further decrease. This is not surprising if we recall the potentials are initialized to the logarithm of the probabilities, $\bTheta=\btheta$, and hence, there may be some class labels which are very unlikely to appear since their score may tend to infinity, and many more iterations may be needed for divMbest to recover $\bmu_u=\0$. Finally, note that the mode accuracy of both versions of divMbest are similar for $\lambda=\eta_u=0.5$, which shows that for small update rates, divMbest can avoid enforcing the constraint $\bmu_u=\0$ by remaining close to the initialization.

\begin{table*}[t]
\caption{{VOC11 segmentation results in the validation set.} We report the performance of the oracle for $20$ hypotheses $(M=20)$. The average score provides the per-class average. We report results for divMbest with $\bmu_u = \0$, $\eta_u=0.5$, and its new version with $\bmu_u = \btheta_u$, $\eta_u=0.5$.
} 
\label{comparisonResultsVOC10}
\centering
\small{
\begin{tabular}{@{\hspace{-1.5mm}}c@{}@{\hspace{0.8mm}}r|*{21}{c@{\hspace{1.5 mm}}} @{\hspace{1.5 mm}}|@{\hspace{1.5 mm}}c@{\hspace{1.5 mm}}|}
    \multicolumn{2}{r|}{}
	&  \begin{sideways}Back. \end{sideways}
 	&  \begin{sideways}Aero. \end{sideways}
 	&  \begin{sideways}Bicy. \end{sideways}
 	&  \begin{sideways}Bird \end{sideways}
 	&  \begin{sideways}Boat \end{sideways}
 	&  \begin{sideways}Bott. \end{sideways}
 	&  \begin{sideways}Bus \end{sideways}
 	&  \begin{sideways}Car \end{sideways}
 	&  \begin{sideways}Cat \end{sideways}
 	&  \begin{sideways}Chair \end{sideways}
 	&  \begin{sideways}Cow \end{sideways}
 	&  \begin{sideways}Table \end{sideways}
 	&  \begin{sideways}Dog \end{sideways}
 	&  \begin{sideways}Horse \end{sideways}
 	&  \begin{sideways}Mbike \end{sideways}
 	&  \begin{sideways}Person \end{sideways}
 	&  \begin{sideways}Plant \end{sideways}
 	&  \begin{sideways}Sheep \end{sideways}
 	&  \begin{sideways}Sofa \end{sideways}
 	&  \begin{sideways}Train \end{sideways}
 	&  \begin{sideways}TV \end{sideways} 	
	&  \begin{sideways}\textbf{Avg.} \end{sideways}\\
\hline

& New Version ($\bmu_u = \btheta_u$) & $87$ & $79$ & $26$ & $62$ & $47$ & $50$ & $70$ & $68$ & $78$ & $21$ & $55$ & $24$ & $61$ & $43$ & $55$ & $54$ & $33$ & $61$ & $31$ & $70$ & $56$ & $53.94$  	\\
  & DivMbest ($\bmu_u = \0$) & $85$ & $73$ & $21$ & $54$ & $40$ & $56$ & $65$ & $66$ & $75$ & $14$ & $50$ & $32$ & $55$ & $41$ & $52$ & $52$ & $36$ & $56$ & $31$ & $63$ & $56$  & $51.12$  	\\

\hline

\end{tabular}}
\end{table*}

\begin{table*}[t!]
\caption{VOC11 segmentation results in the validation set with missing potentials. We report the performance of the oracle for $20$ hypotheses $(M=20)$ with $2\%$ and $10\%$ of observed potentials, and the rest set as missing. The average score provides the per-class average. We report results for {divMbest} with $\bmu_u=\0$, $\eta_u=5$, and its new version with $\bmu=\btheta$, $\eta_u=5$, $\eta_p=0.25$.
} 
\label{comparisonResultsVOC10MissUn}
\centering
\small{
\begin{tabular}{@{\hspace{-1.5mm}}c@{}@{\hspace{0.8mm}}r|*{21}{c@{\hspace{1.5 mm}}} @{\hspace{1.5 mm}}|@{\hspace{1.5 mm}}c@{\hspace{1.5 mm}}|}
    \multicolumn{2}{r|}{}
	&  \begin{sideways}Back. \end{sideways}
 	&  \begin{sideways}Aero. \end{sideways}
 	&  \begin{sideways}Bicy. \end{sideways}
 	&  \begin{sideways}Bird \end{sideways}
 	&  \begin{sideways}Boat \end{sideways}
 	&  \begin{sideways}Bott. \end{sideways}
 	&  \begin{sideways}Bus \end{sideways}
 	&  \begin{sideways}Car \end{sideways}
 	&  \begin{sideways}Cat \end{sideways}
 	&  \begin{sideways}Chair \end{sideways}
 	&  \begin{sideways}Cow \end{sideways}
 	&  \begin{sideways}Table \end{sideways}
 	&  \begin{sideways}Dog \end{sideways}
 	&  \begin{sideways}Horse \end{sideways}
 	&  \begin{sideways}Mbike \end{sideways}
 	&  \begin{sideways}Person \end{sideways}
 	&  \begin{sideways}Plant \end{sideways}
 	&  \begin{sideways}Sheep \end{sideways}
 	&  \begin{sideways}Sofa \end{sideways}
 	&  \begin{sideways}Train \end{sideways}
 	&  \begin{sideways}TV \end{sideways} 	
	&  \begin{sideways}\textbf{Avg.} \end{sideways}\\
\hline

&{ New Version ($\bmu=\btheta$) \hfill $2\%$} & $78$ & $6$ & $3$ & $35$ & $50$ & $55$ & $57$ & $54$ & $60$ & $34$ & $55$  & $64$  & $52$  & $45$  & $45$  & $49$  & $42$  & $54$  & $60$  & $58$  & $59$  & $48.33$  	\\
  &{ DivMbest ($\bmu_u=\0$) \hfill $2\%$} & $73$ & $13$ & $6$ & $15$ & $11$ & $31$ & $33$ & $31$ & $29$ & $16$ & $38$  & $42$  & $27$  & $19$  & $25$  & $28$  & $19$  & $23$  & $40$  & $30$  & $34$  & $27.77$ 	\\
\hline
\hline
&{ New Version ($\bmu=\btheta$) \hfill  $10\%$} & $86$ & $13$ & $10$ & $68$ & $72$ & $72$ & $74$ & $70$ & $78$ & $51$ & $75$  & $74$  & $74$  & $69$  & $61$  & $67$  & $57$  & $73$  & $78$  & $75$  & $76$  & $65.36$  	\\
  &{ DivMbest ($\bmu_u=\0$) \hfill $10\%$} & $81$ & $27$ & $7$ & $24$ & $32$ & $46$ & $63$ & $50$ & $58$ & $31$ & $57$  & $62$  & $50$  & $44$  & $49$  & $48$  & $42$  & $41$  & $60$  & $57$  & $65$  & $47.37$   	\\
\hline
\hline
&{ New Version ($\bmu=\btheta$) \hfill $100\%$} & $99$ & $96$ & $74$ & $98$ & $94$ & $98$ & $97$ & $95$ & $98$ & $93$ & $98$  & $97$  & $98$  & $97$  & $96$  & $97$  & $92$  & $98$  & $98$  & $97$  & $96$  & $95.53$  	\\
  &{ DivMbest ($\bmu_u=\0$) \hfill $100\%$} & $99$ & $96$ & $74$ & $98$ & $94$ & $98$ & $97$ & $95$ & $98$ & $93$ & $98$  & $97$  & $98$  & $97$  & $96$  & $97$  & $92$  & $98$  & $98$  & $97$  & $96$  & $95.51$   	\\
\hline
\end{tabular}
}
\end{table*}

\begin{figure*}[t!]
 \centering
\begin{tabular}{@{\hspace{-0.6 mm}}c@{\hspace{0.2 mm}}c@{\hspace{1.7 mm}}c@{\hspace{0.6 mm}}c@{\hspace{0.6 mm}}c@{\hspace{2 mm}}c@{\hspace{0.6 mm}}c@{\hspace{0.6 mm}}c}

\
&&\multicolumn{3}{c}{\hspace*{-2.5mm}\footnotesize{\bf{--- --- New Version ($\bmu_u=\btheta_u$)} --- ---}}&\multicolumn{3}{c}{\footnotesize{\bf{--- --- DivMbest ($\bmu_u=\0$)} --- ---}}\\

\vspace{-0.5 mm}
\footnotesize{\bf{Original}}&\footnotesize{\bf{Ground-truth}}&\footnotesize{\bf{Hypothesis}}&\footnotesize{\bf{Hypothesis}}&\footnotesize{\bf{Mode}}&\footnotesize{\bf{Hypothesis}}&\footnotesize{\bf{Hypothesis}}&\footnotesize{\bf{Mode}}\\

\vspace{-0.5 mm}
 \includegraphics[width=0.11\textwidth]{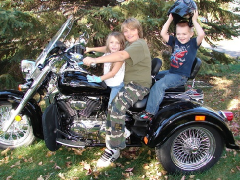} &
 \includegraphics[width = 0.11\textwidth]{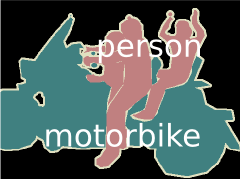} &
 \includegraphics[width = 0.11\textwidth]{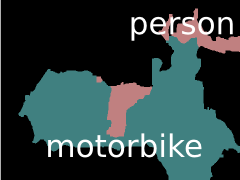} &
 \includegraphics[width = 0.11\textwidth]{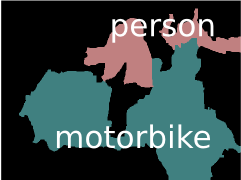} & 
 \includegraphics[width = 0.11\textwidth]{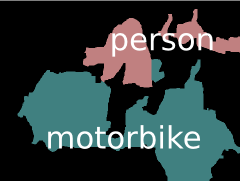}&
 \includegraphics[width = 0.11\textwidth]{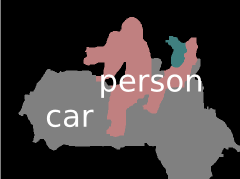}&
 \includegraphics[width = 0.11\textwidth]{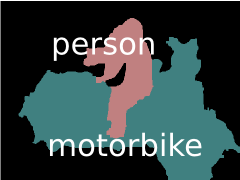}&
 \includegraphics[width = 0.11\textwidth]{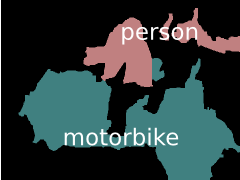} \\

\vspace{-0.5 mm}
  \includegraphics[width=0.11\textwidth]{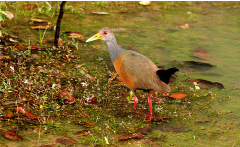} &
 \includegraphics[width = 0.11\textwidth]{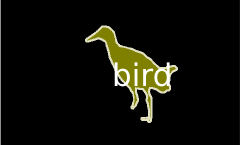} &
 \includegraphics[width = 0.11\textwidth]{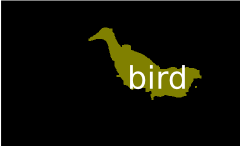} &
 \includegraphics[width = 0.11\textwidth]{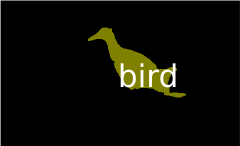} & 
 \includegraphics[width = 0.11\textwidth]{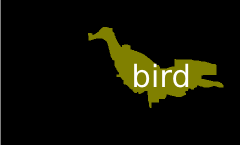}&
 \includegraphics[width = 0.11\textwidth]{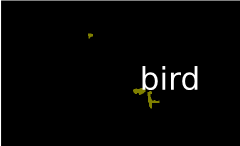}&
 \includegraphics[width = 0.11\textwidth]{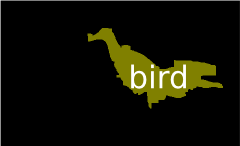}&
 \includegraphics[width = 0.11\textwidth]{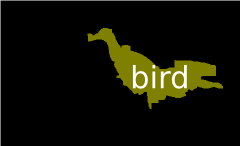} \\

&&\multicolumn{3}{c}{\hspace*{-2.5mm}\footnotesize{\bf{--- --- New Version ($\bmu=\btheta)$} --- }}&\multicolumn{3}{c}{\footnotesize{\bf{--- --- DivMbest ($\bmu_u=\0$)} --- ---}}\\

\footnotesize{\bf{Original}}& \footnotesize{\bf{Ground-truth}}&\multicolumn{3}{c}{\hspace*{-2.5mm}\footnotesize{\bf{Hypotheses ($2\%$ known unary terms)} }} &\multicolumn{3}{c}{\hspace*{-2.5mm}\footnotesize{\bf{ Hypotheses ($2\%$ known unary terms)} }}\\

\vspace{-0.5 mm}
 \includegraphics[width=0.11\textwidth]{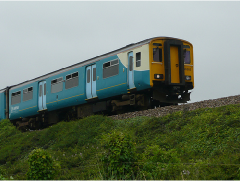} &
 \includegraphics[width = 0.11\textwidth]{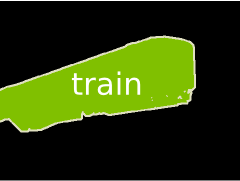} &
 \multicolumn{3}{l}{\hspace*{-0.2cm}
 \includegraphics[width = 0.11\textwidth]{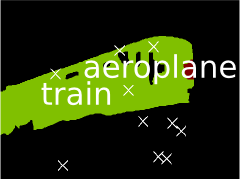} 
 \includegraphics[width = 0.11\textwidth]{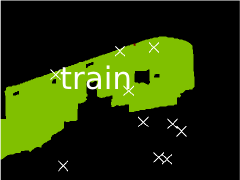}  
 \includegraphics[width = 0.11\textwidth]{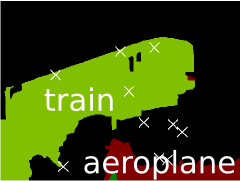}}&
  \multicolumn{3}{l}{\hspace*{-0.2cm}
 \includegraphics[width = 0.11\textwidth]{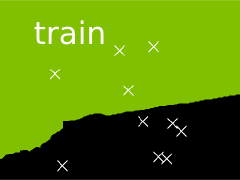}
 \includegraphics[width = 0.11\textwidth]{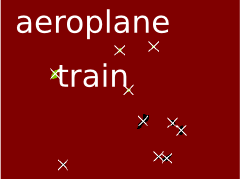}
 \includegraphics[width = 0.11\textwidth]{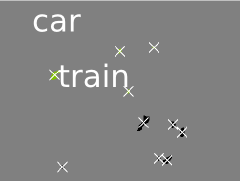}} \\

\vspace{-0.5 mm}
 \includegraphics[width=0.11\textwidth]{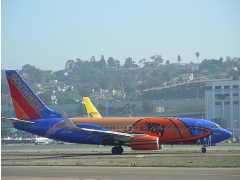} &
 \includegraphics[width = 0.11\textwidth]{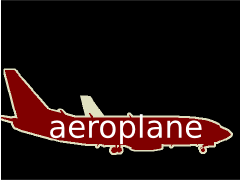} &
 \multicolumn{3}{l}{\hspace*{-0.2cm}
 \includegraphics[width = 0.11\textwidth]{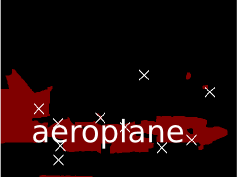} 
 \includegraphics[width = 0.11\textwidth]{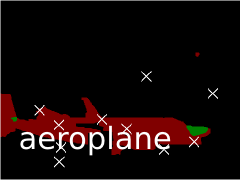}  
 \includegraphics[width = 0.11\textwidth]{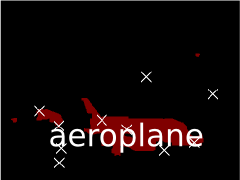}}&
  \multicolumn{3}{l}{\hspace*{-0.2cm}
 \includegraphics[width = 0.11\textwidth]{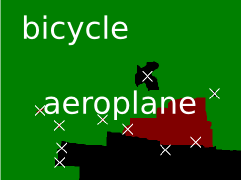}
 \includegraphics[width = 0.11\textwidth]{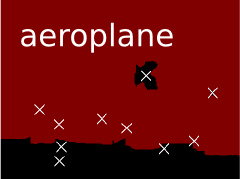}
 \includegraphics[width = 0.11\textwidth]{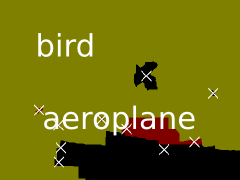}} \\

\end{tabular}
\caption{\emph{Examples of segmented images on VOC11.} Hypotheses in the first and second rows are obtained from divMbest, and its new version with $\bmu_u=\btheta_u$. Mode is computed with $20$ hypotheses ($M=20$). Hypotheses in the third and fourth rows are obtained from {divMbest} and its new version with $\bmu=\btheta$ with $2\%$ observed unary potentials (indicated with white crosses in the image). 
}
\label{figimHerdingRobustExample}
\end{figure*}

In Fig.~\ref{fig:herdingWithVsWithoutExt}b, we report the oracle accuracy for both versions of divMbest, for the update rate $\lambda=\eta_u=0.5$ (these are the best parameters found during the cross-validation for both versions of divMbest, we did not found better results with $\lambda$ smaller than $0.5$), and in Table~\ref{comparisonResultsVOC10} we report their per-class accuracy for $M=20$.
We can observe that the version of divMbest that does not enforce equiprobable marginals outperforms divMbest. This is because our version of divMbest encourages to appear more often the object categories that are more likely, rather than making all object categories equiprobable.

\subsection{Results on Interactive Segmentation}
In this setup, divMbest can not exploit using small update rates to remain close to the initialization, because the initialization is not useful anymore, as we show next. This serves us to further demonstrate the inconvenience of the constraint  $\bmu_u=\0$, and how it can be alleviated.

Recall that divMbest forces all the unary terms to recover the moments $\bmu_u=\0$, including the  unary terms that we have no information in the interactive segmentation. Thus, all unary terms, both known and unknown, have to be initialized in order to be updated. We initialize all the unknown unary terms to a predefined value, which is equal to $0$.
Note that now it is not only the constraint $\bmu_u=\0$ that is not representative of the input image, but also the initialization of the unknown unary terms.
Thus, using a small update rate to remain close the the initialization could not be effective anymore. 
This allows for a controlled experiment in which the amount of unknown unary potentials is related to the quality of the initialization.

We evaluate divMbest, and also, the new version of divMbest that we introduced before for semantic segmentation,~\ie the new version of divMbest that uses $\bmu_u=\tilde{\btheta}_u$. Note that this version of divMbest  does not need to update the unknown unary potentials as in divMbest, because the unary moments only constrain the known unary potentials. Thus,  the new version does not suffer as much as divMbest from the initialization problem.
However, since in the interactive segmentation the unary potentials are the ground-truth indicated by the user, the new version of divMbest does not produce diverse samples, as the unary moments encourage that $100\%$ of the samples take the value of the ground-truth. To avoid this, we also use the pairwise moments in the new version of divMbest, and we set these moments in the same way as the unary moments: the pairwise moments are equal to the pairwise potential  (Potts potential) normalized to sum $1$ in order to be a marginal probability. Note that this does not add  any additional information to the model, as it uses the same CRF as before, and it only changes the constraints of the moments. Thus, in the new version of divMbest that we use in this experiment, the unary moments do not enforce equiprobable labels, and the pairwise moments allow to extract multiple hypotheses. We call this version of divMbest using $\bmu=\tilde{\btheta}$ because it uses both unary and pairwise moments based on the information in the potentials of divMbest.

In Fig.~\ref{fig:herdingWithVsWithoutExt}c we report the accuracy for the oracle criterion for the interactive image segmentation application with $20$ number of hypotheses, $M=20$. In Table~\ref{comparisonResultsVOC10MissUn} we report the per-class accuracy for this experiment. The update rates are set by cross-validation to $\lambda=\eta_u=0.15$ for divMbest and to  $\eta_u=0.75$  and the pairwise update rate to $\eta_p = 0.25$ for the divMbest with $\bmu=\tilde{\btheta}$. We can see that with $2\%$ of observed unary terms, the new version of divMbest substantially improves the accuracy of {divMbest}, more than $15\%$. This shows that when the initialization of {divMbest} is implausible,  even if we use a small update rate, the constraints enforced by divMbest make the dynamical system easily go astray. When more unary terms are observed, the performance of divMbest recovers and becomes similar as in the semantic segmentation experiment in Fig.~\ref{fig:herdingWithVsWithoutExt}a. This demonstrates that not always the implausible constraint $\bmu_u=\0$ can be alleviated by using small update rates to remain close to the initialization.

{\bf Qualitative Examples}
 In Fig.~\ref{figimHerdingRobustExample} some examples are depicted. We can see that the difference between the hypotheses from divMbest have a more varied set of class labels compared to the new version of divMbest, which respect more the output of the classifier. This is in accordance to the observation that {divMbest} enforces the recovery of the unary moments equal to zero. 

\section{Conclusions}

We showed that Herding generalizes the divMbest algorithm by~\cite{batra2012diverse},~\ie
divMbest is an instance of Herding with ${\bmu_u=\0}$, ${\eta_p=0}$, and $\bTheta=\btheta$. We have shown that these constraints are implausible in practice, as they enforce equiprobable marginals. In practice, this can be alleviated by using a small update rate to remain close to the initialization.
Also, we have shown that there are applications where the implausible constraints can be replaced by other constraints that produce significant improvements of the accuracy when a good initialization is not available.
We expect that in the future this generalization can bring Herding to other applications in which multiple hypotheses are used.


\bibliographystyle{ieeetr}
\bibliography{egbib}

\vspace*{-1cm}
 

\begin{IEEEbiographynophoto}{Ece Ozkan} received her B.Sc. and M.Sc. degrees in electrical engineering at ETH Zurich, Switzerland in 2012 and 2014, respectively. She  is  currently  a doctoral student in Computer-assisted applications in Medicine Group at ETH Zurich. Her  research  interests  include  elastography, image-
guided therapy, beamforming, and motion estimation.
\end{IEEEbiographynophoto}

\vspace*{-1cm}
\begin{IEEEbiographynophoto}{Gemma Roig} pursued her PhD in Computer Vision at ETH Zurich. She is currently a postdoc at MIT in the Center for Brains Minds and Machines. She is also affiliated at the Laboratory for Computational and Statistical Learning, which is a collaborative agreement between the Istituto Italiano di Tecnologia and MIT. Her research interests include computational models of human vision, artificial intelligence and computer vision. 
\end{IEEEbiographynophoto}

\vspace*{-1cm}
\begin{IEEEbiographynophoto}{Orcun Goksel}
 received two BSc degrees in electrical engineering (2001) and in computer science (2002) from Middle East Technical University, Ankara, Turkey. He received his MASc (2004) and PhD (2009) degrees in Electrical and Computer Engineering at the University of British Columbia, Vancouver, Canada. Following postdoc and senior scientist positions, since 2014 he is an assistant professor at ETH Zurich, Switzerland. He has received the 2016 ETH Spark Award, the 2014 CTI Swiss MedTech Award, and the 2011 WAGS Innovation in Technology Award. His research interests include ultrasound imaging, medical image analysis, tissue biomechanical characterization, patient-specific modelling, image-guided therapy, and medical simulation in virtual-reality.
\end{IEEEbiographynophoto}

\vspace*{-1cm}
\begin{IEEEbiographynophoto}{Xavier Boix} obtained his PhD at ETH Zurich in 2014. During his doctoral thesis work, he received the European Intel Doctoral Student Award (2013) and won the PASCAL VOC Challenge in semantic segmentation (2010).  Afterwards, he was a postdoc at NUS Singapore, frequenting MIT as an affiliate until 2016. Now he is a postdoc at MIT in the Center for Brains Minds and Machines, and he is also affiliated at the Laboratory for Computational and Statistical Learning at the Istituto Italiano di Tecnologia. His research interests include computational vision and neuroscience, machine learning and computer vision.
\end{IEEEbiographynophoto}

\vfill

\enlargethispage{-5in}

\end{document}